# Batch-iFDD for Representation Expansion in Large MDPs

Alborz Geramifard[†]    Thomas J. Walsh[†]    Nicholas Roy[⋆]    Jonathan P. How[†]

[†]Laboratory for Information and Decision Systems
[⋆]Computer Science and Artificial Intelligence Laboratory
Massachusetts Institute of Technology
77 Massachusetts Ave., Cambridge, MA 02139

## Abstract

Matching pursuit (MP) methods are a promising class of feature construction algorithms for value function approximation. Yet existing MP methods require creating a pool of potential features, mandating expert knowledge or enumeration of a large feature pool, both of which hinder scalability. This paper introduces batch incremental feature dependency discovery (Batch-iFDD) as an MP method that inherits a provable convergence property. Additionally, Batch-iFDD does not require a large pool of features, leading to lower computational complexity. Empirical policy evaluation results across three domains with up to one million states highlight the scalability of Batch-iFDD over the previous state of the art MP algorithm.

## 1 Introduction

In complex decision-making tasks, from stacking blocks to flying surveillance missions, the number of possible features used to represent a domain grows exponentially in the basic number of variables. It follows that generating a small number of relevant features that are sufficient for determining an optimal policy is a critical component for tractable reinforcement learning in complex environments. However, even in the well-studied case of linear value function approximation [Silver *et al.*, 2008; Stone *et al.*, 2005], finding the "right" set of features remains a challenge.

In the linear value function approximation case, several methods exist for automated feature construction [*e.g.,* Lin and Wright, 2010; Ratitch and Precup, 2004; Whiteson *et al.*, 2007]. Of these techniques, Matching Pursuit (MP) algorithms [Painter-Wakefield and Parr, 2012] have shown great promise in incrementally expanding the set of features to better model the value function. However, all prior MP techniques begin with a collection of potential features from which new features are selected. In large MDPs, this pool of features has to either be carefully selected by a domain expert or be prohibitively large to include all critical features, both hindering scalability. Still, despite such requirements, MP algorithms have various desirable properties [See Painter-Wakefield and Parr, 2012], making them an attractive option for RL feature construction.

Some similar techniques avoid enumerating a set of potential features, but are not scalable for other reasons. For example, Bellman Error Basis Function (BEBF) [Parr *et al.*, 2007] iteratively constructs a set of basis vectors without an enumerated pool of potential features. However, BEBF relies on supervised learning techniques to map states to their feature values. This process can be as complex as determining the value function itself, mitigating the tractability gains of feature construction. Similarly, Proto-Value Functions [Mahadevan *et al.*, 2006] do not use a pool of potential features but learn a complex manifold representation that can be computationally intensive for arbitrary MDPs.

This paper presents a new algorithm, Batch incremental Feature Dependency Discovery (Batch-iFDD), that does not require a large set of potential features at initialization. Moreover, we prove Batch-iFDD is an MP algorithm, thereby inheriting the theoretical benefits associated with those techniques. Batch-iFDD extends the previously described online incremental Feature Dependency Discovery (iFDD) algorithm [Geramifard *et al.*, 2011], which creates increasingly finer features that help to eliminate error of the value function approximation. Our contributions in this paper are to (1) extend iFDD to the batch setting (Section 2.4), (2) prove that Batch-iFDD is an MP algorithm (Corollary 3.6) and derive its guaranteed rate of error-bound reduction (Theorem 3.4), (3) derive a practical approximation for iFDD's objective function resulting in an algorithm called Batch-iFDD[+] (Equation 10), and (4) empirically compare Batch-iFDD with the state of the art MP algorithm across three domains including a 20 dimensional version of the System Administrator domain with over one million states (Section 4).

## 2 Preliminaries

In this section we define data structures for modeling RL domains and approximating value functions. We also describe a basic reinforcement learning technique (Temporal Difference Learning) for evaluating a policy's value through experience. Finally, we provide definitions of the relationships between features and describe the feature search process.

### 2.1 Reinforcement Learning

A **Markov Decision Process (MDP)** is a tuple $(\mathcal{S}, \mathcal{A}, \mathcal{P}_{ss'}^a, \mathcal{R}_{ss'}^a, \gamma)$ where $\mathcal{S}$ is a set of states, $\mathcal{A}$ is a set of actions, $\mathcal{P}_{ss'}^a$ is the probability of getting to state $s'$ by taking action $a$ in state $s$, $\mathcal{R}_{ss'}^a$ is the corresponding reward, and $\gamma \in [0, 1)$ is a discount factor that balances current and future rewards. We focus on MDPs with finite states. A *trajectory* is a sequence $s_0, a_0, r_0, s_1, a_1, r_1, s_2, \ldots$, where the action $a_t \in \mathcal{A}$ is chosen according to a deterministic *policy* $\pi: \mathcal{S} \to \mathcal{A}$, mapping each state to an action. Given a policy $\pi$, the value function, $V^\pi(s)$ for each state, is the expected sum of the discounted rewards for an agent starting at state $s$ and then following policy $\pi$ thereafter:

$$\begin{aligned} V^\pi(s) &= E_\pi \left[ \sum_{t=0}^\infty \gamma^t r_t \Big| s_0 = s \right] \\ &= \sum_{s' \in \mathcal{S}} \mathcal{P}_{ss'}^{\pi(s)} \left[ \mathcal{R}_{ss'}^{\pi(s)} + \gamma V^\pi(s') \right]. \end{aligned}$$

Since this paper primarily addresses the policy evaluation problem (*i.e.,* finding the value function of a fixed policy), the $\pi$ notation is dropped from this point on and included implicitly. For a finite-state MDP, the vector $\boldsymbol{V}_{|\mathcal{S}| \times 1}$ represents the value function. The matrix $\boldsymbol{P}_{|\mathcal{S}| \times |\mathcal{S}|}$ represents the transition model with $\boldsymbol{P}_{ij} = \mathcal{P}_{s_i s_j}^{\pi(s_i)}$, and the vector $\boldsymbol{R}_{|\mathcal{S}| \times 1}$ is the reward model, with $\boldsymbol{R}_i = \sum_j \mathcal{P}_{s_i s_j}^{\pi(s_i)} \mathcal{R}_{s_i s_j}^{\pi(s_i)}$. Hence $\boldsymbol{V}$ can be calculated in the matrix form as $\boldsymbol{V} = \boldsymbol{R} + \gamma \boldsymbol{P} \boldsymbol{V} = \mathbf{T}(\boldsymbol{V})$, where $\mathbf{T}$ is the Bellman operator.

Storing a unique value for each state is impractical for large state spaces. A common approach is to use a linear approximation of the form $V(s) = \boldsymbol{\theta}^\top \phi(s)$. The feature function $\phi: \mathcal{S} \to \mathbb{R}^n$ maps each state to a vector of scalar values. Each element of the feature function $\phi(s)$ is called a *feature*; $\phi_f(s) = c \in \mathbb{R}$ denotes that feature $f$ has scalar value $c$ for state $s$, where $f \in \chi = \{1, \ldots, n\}$. $\chi$ represents the set of features;[1]

---
[1]Properties such as being close to a wall or having low

the vector $\boldsymbol{\theta} \in \mathbb{R}^n$ holds weights. Hence,

$$\boldsymbol{V} \approx \tilde{\boldsymbol{V}} = \begin{bmatrix} \text{---} \phi^\top(s_1) \text{---} \\ \text{---} \phi^\top(s_2) \text{---} \\ \vdots \\ \text{---} \phi^\top(s_{|\mathcal{S}|}) \text{---} \end{bmatrix} \times \begin{bmatrix} \theta_1 \\ \theta_2 \\ \vdots \\ \theta_n \end{bmatrix} \triangleq \boldsymbol{\Phi} \boldsymbol{\theta}.$$

Binary features ($\phi_f : \mathcal{S} \to \{0, 1\}$) are of special interest to practitioners [Silver *et al.*, 2008; Stone *et al.*, 2005; Sturtevant and White, 2006], mainly because they are computationally cheap, and are the focus of this paper.

The **Temporal Difference Learning (TD)** [Sutton, 1988] algorithm is a traditional policy evaluation method where the current $V(s)$ estimate is adjusted based on the difference between the current estimated state value and a better approximation formed by the actual observed reward and the estimated value of the following state. Given $(s_t, r_t, s_{t+1})$ and the current value estimates, the TD-error, $\delta_t$, is calculated as: $\delta_t(V) = r_t + \gamma V(s_{t+1}) - V(s_t)$. The one-step TD algorithm, also known as TD(0), updates the value estimates using $V(s_t) = V(s_t) + \alpha \delta_t(V)$, where $\alpha$ is the learning rate. In the case of linear function approximation, the TD update can be used to change the weights of the value function approximator: $\boldsymbol{\theta}_{t+1} = \boldsymbol{\theta}_t + \alpha \delta_t(V)$. In the batch setting, the least-squares TD (LSTD) algorithm [Bradtke and Barto, 1996] finds the weight vector directly by minimizing the sum of TD updates over all the observed data.

### 2.2 Matching Pursuit Algorithms

This paper considers algorithms that expand features during the learning process. The class of matching pursuit (MP) algorithms, such as OMP-TD [Painter-Wakefield and Parr, 2012] has been shown recently to be a promising approach for feature expansion. An algorithm is MP if it selects the new feature from the pool of features that has the highest correlation with the residual.

### 2.3 Finer (Coarser) Features and Search

We now define some properties of state features and feature-search algorithms. The *coverage* of a feature is the portion of the state space for which the feature value is active (*i.e.,* non-zero). We say that a feature $A$ is *coarser* than feature $B$ ($B$ is *finer* than $A$) if $A$ has a higher coverage than $B$. For example, consider a task of administrating 3 computers ($C_1, C_2$ and $C_3$) that each can be up or down. Feature $A$ ($C_1 =$ down) is coarser than feature $B$ ($C_3 =$ down AND $C_2 =$ up), because coverage(A) = 0.5 > coverage(B) = 0.25.

---
fuel can be considered as features. In our setting, we assume all such properties are labeled with numbers and are addressed with their corresponding number.

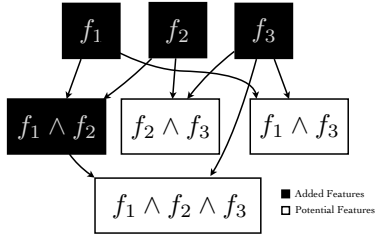

Figure 1: A partial concept lattice showing potential features as identified by iFDD. With methods like OMP-TD, all nodes are considered as potential features, despite their location in the lattice.

High coverage is not always the best criterion for selecting features because the resulting partitions may not yield a good approximation of the value function. For example, suppose that having $C_1 =$ down translates into negative values except when $C_3 =$ up. If the weight corresponding to feature $A$ ($C_1 =$ down) is set to a negative value, the value function is reduced for all parts of the state space covered by $A$, including situations where $C_3 =$ up. The problem of feature construction is to find a small set of features with high coverage that still approximate the value function with bounded error.

One approach is to assume a set of *base* binary features that, in full combination, uniquely describe each state. These features would have high coverage, and constitute a very large set. Combining base features with the conjunction operator would allow us to find features with lower coverage. This search process can be thought of as selecting nodes in a graph structure (Figure 1). Each node represents a feature, with the top level nodes corresponding to the domain's base features. An edge indicates the origin feature subsumes the destination feature (and therefore the destination is finer). For instance, in our example, $((C_1 = $ up$) \to (C_1 = $ up $\land C_2 = $ up$))$ would be an edge.

Current MP methods such as OMP-TD require an enumerated set of potential features (*e.g.,* all possible feature conjunctions of base features) that can be combinatorially large. MP methods iterate over this set on every step of feature generation, leading to high computational demand if the set is large, or poor performance if only a few nodes in the lattice are considered. Methods such as BEBF [Parr *et al.*, 2007] adopt an alternative approach by creating new features using supervised learning that often do not have a clean logical interpretation and can be arbitrary complex. Hence, this paper focuses on mapping states to conjunctive features in the concept lattice. Ideally, a search method would find relevant features in the lattice by selectively growing the tree as opposed to current MP techniques that have to be initialized with the set of all potential features.

### 2.4 iFDD and Batch-iFDD

The iFDD algorithm [Geramifard *et al.*, 2011] is an online feature expansion technique with low computational complexity. Given an initial set of base features, iFDD adds the conjunction of existing binary features as new features. Following the original work, we restrict new features to be the conjunction of previously selected features, which still gives us a rich set of potential features without introducing complex reasoning into the search process. At each time-step, iFDD performs the following steps:

**1.** Identify pair-wise combinations of active features.
**2.** Accumulate the absolute value of the observed TD-error for all such pairs.
**3.** If the accumulated value for a pair of features $f$ and $g$ exceeds a predefined threshold, add feature $f \land g$ to the pool of features.

Within the concept lattice described in Section 2.3, the first step considers features where two parent concepts are already in the feature space, and the conjunction of these parents is equivalent to the potential feature. Then potential features that also reduce the value function approximation error significantly are added to the feature set. This algorithm has the ability to include fine-grained features where they are necessary to represent the value function, but can avoid other (less helpful) features of similar complexity.

This paper uses iFDD for policy evaluation in a batch setting where LSTD estimates the TD-error over all samples and then the most "relevant" feature is added to the features. We now analyze the behavior of Batch-iFDD and through this analysis derive a new algorithm, Batch-iFDD$^+$, which better approximates the best guaranteed rate of error-bound reduction.

## 3 Theoretical Results

Geramifard *et al.* [2011] introduced iFDD, empirically verified the approach, and proved the asymptotic convergence of iFDD combined with TD. This work extends those theoretical results by showing that executing iFDD combined with TD in the batch setting is equivalent to approximately finding the feature from the conjunction of existing features with the maximum guaranteed error-bound reduction.

In order to use the conjunction operator to define new features, we redefine the notion of a feature. Given an initial feature function $\phi$ outputting vectors in $\mathbb{R}^n$, we address each of its $n$ output elements as an *index* rather than a *feature* from this point on; $\phi_i(s) = c$ denotes index $i$ of the initial feature function has value $c$ in state $s$.[2] The set of all indices is $\mathcal{V}_n = \{1, \cdots, n\}$. A feature, $f$, is redefined as an arbitrary subset of $\mathcal{V}_n$, where $\phi_f(s) \triangleq \bigwedge_{i \in f} \phi_i(s)$, and

$$\phi_{\{\}}(s) \triangleq \begin{cases} 1 & \text{if } \forall i \in \mathcal{V}_n, \phi_i(s) = 0 \\ 0 & \text{otherwise} \end{cases},$$

where $\{\}$ is the null feature. For example $\phi_{\{1,3\}}(.) = \phi_1(.) \wedge \phi_3(.)$. Notice that a single index can constitute a feature (e.g., $f = \{1\}$). Further we assume that all sets are ordered based on the cardinality size of each element in ascending order, unless specified. Given a set of features $\chi = \{f_1, f_2, ...f_N\}$, the definition of $\Phi$ is extended as follows:

$$\Phi_\chi = \begin{bmatrix} \phi_{f_1}(s_1) & \phi_{f_2}(s_1) & \cdots & \phi_{f_N}(s_1) \\ \phi_{f_1}(s_2) & \phi_{f_2}(s_2) & \cdots & \phi_{f_N}(s_2) \\ \vdots & \vdots & \ddots & \vdots \\ \phi_{f_1}(s_{|\mathcal{S}|}) & \phi_{f_2}(s_{|\mathcal{S}|}) & \cdots & \phi_{f_N}(s_{|\mathcal{S}|}) \end{bmatrix}_{|\mathcal{S}| \times N}.$$

For brevity, we define $\boldsymbol{\phi}_f$ as a column of feature matrix $\Phi_\chi$ that corresponds to feature $f$. $\boldsymbol{\phi}_f = \Phi_{\{f\}} = \begin{bmatrix} \phi_f(s_1) & \phi_f(s_2) & \cdots & \phi_f(s_m) \end{bmatrix}^\top$. Also, define $B_n \triangleq \{\{\}, \{1\}, \{2\}, \cdots, \{n\}\}$, as the set of features with cardinality less than 2, and $\mathcal{F}_n \triangleq \wp(\mathcal{V}_n)$ as the set of all possible features. $\wp$ is the power set function (i.e., the function that returns the set of all possible subsets). Hence, $\chi \subseteq \mathcal{F}_n$ is an arbitrary set of features. Further, define operator $\texttt{pair} : \chi_1 \to \chi_2$, where $\chi_1, \chi_2 \subseteq \mathcal{F}_n$:

$$\texttt{pair}(\chi) \triangleq \left\{ f \cup g \,\middle|\, f, g \in \chi, f \cup g \notin \chi \right\},$$
$$\texttt{pair}^k(\chi) \triangleq \underbrace{\texttt{pair}(\texttt{pair}(\cdots(\texttt{pair}(\chi))))}_{k \text{ times}},$$

$\texttt{pair}^0(\chi) \triangleq \chi, \texttt{full}(\chi) \triangleq \bigcup_{i=0,\cdots,n} \texttt{pair}^i(\chi)$. Essentially, the $\texttt{pair}$ operator provides the set of all possible new features built on the top of a given set of features using pairwise conjunction. The $\texttt{full}$ operator generates all possible features given a set of features.

Now, given an MDP with a fixed policy, the feature expansion problem can be formulated mathematically. Given $\chi$ as a set of features and the corresponding approximation of the value function under the fixed policy, $\tilde{V}_\chi = \Phi_\chi \theta$, find $f \in \texttt{pair}(\chi)$ that maximizes the following error reduction:

$$ER = \|V - \tilde{V}_\chi\| - \|V - \tilde{V}_{\chi \cup \{f\}}\|, \quad (1)$$

where $\|.\|$ is the $\ell_2$ norm weighted by the steady state distribution. Consequently, all our theoretical analyses are performed in the weighted Euclidean space following the work of [Parr *et al.*, 2007]. The theorem and proof following the next set of assumptions provide an analytical solution that maximizes Equation 1.

---
[2]Note switch in subscript from $f$ (feature) to $i$ (index).

**Assumptions**:
**A1.** The MDP has a binary $d$-dimensional state space, $d \in \mathbb{N}^+$ (i.e., $|\mathcal{S}| = 2^d$). Furthermore, each vertex in this binary space corresponds to one unique state; $s \in \{0, 1\}^d$.
**A2.** The agent's policy, $\pi$, is fixed.
**A3.** Each initial feature corresponds to a coordinate of the state space (i.e., $\phi(s) = s$). Hence the number of indices, $n$, is equal to the number of dimensions, $d$.

Assumption A1 is a more specific form of a general assumption where each dimension of the MDP can be represented as a finite vector and each dimension has a finite number of possible values. It is simple to verify that such an MDP can be transformed into an MDP with binary dimensions. This can be done by transforming each dimension of the state space with $M$ possible values into $M$ binary dimensions. The MDP with binary dimensions was considered for brevity of the proofs.

**Definition** The angle between two vectors $\boldsymbol{X}, \boldsymbol{Y} \in \mathbb{R}^d, d \in \mathbb{N}^+$ is the smaller angle between the lines formed by the two vectors: $\angle(\boldsymbol{X}, \boldsymbol{Y}) = \arccos\left(\frac{|\langle \boldsymbol{X} \cdot \boldsymbol{Y}\rangle|}{\|\boldsymbol{X}\| \cdot \|\boldsymbol{Y}\|}\right)$, where $\langle \cdot, \cdot \rangle$ is the weighted inner product operator. Note that $0 \leq \angle(\boldsymbol{X}, \boldsymbol{Y}) \leq \frac{\pi}{2}$.

**Theorem 3.1** *Given Assumptions A1-A3 and a set of features $\chi$, where $B_n \subseteq \chi \subseteq \mathcal{F}_n$, then feature $f^* \in \Omega = \{f | f \in \texttt{pair}(\chi), \angle(\boldsymbol{\phi}_f, \boldsymbol{\delta}) < \arccos(\gamma)\}$ with the maximum guaranteed error-bound reduction defined in Equation 1 can be calculated as:*

$$f^* = \operatorname*{argmax}_{f \in \Omega} \frac{|\sum_{s \in \mathcal{S}, \phi_f(s)=1} \mathbf{d}(s)\boldsymbol{\delta}(s)|}{\sqrt{\sum_{s \in \mathcal{S}, \phi_f(s)=1} \mathbf{d}(s)}}, \quad (2)$$

*where $\boldsymbol{\delta} = \mathbf{T}(\tilde{V}_\chi) - \tilde{V}_\chi$ is the Bellman error vector, and $\mathbf{d}$ is the steady state distribution vector.*

The rest of this section provides the building blocks of the proof, followed by a discussion of the theoretical result. Theorem 3.2 states that given an initial set of features, the feature matrix is always full column rank through the process of adding new features using the $\texttt{pair}$ operator. Lemma 3.3 provides a geometric property for vectors in $d$-dimensional space under certain conditions. Theorem 3.4 provides a general guaranteed rate of error-bound reduction when adding arbitrary features to the representation in addition to the convergence proof stated in Theorem 3.6 of [Parr *et al.*, 2007]. Theorem 3.5 narrows down Theorem 3.4 to the case of binary features, where new features are built using the `pair` operator. Finally Theorem 3.1, as stated above, concludes that given the set of potential features obtained by the `pair` operator and filtered based on the their angle with the Bellman error vector; the one with the maximum guaranteed error-bound reduction is identified by Equation 2.

**Theorem 3.2** *Given Assumptions A1-A3, $\forall \chi \subseteq \mathcal{F}_n, \Phi_\chi$ has full rank.*

**Proof** In appendix.

**Insight:** Theorem 3.2 shows that the conjunction operator creates a matrix $\Phi_{\mathcal{F}_n}$ that forms a basis for $\mathbb{R}^{|\mathcal{S}|}$ (*i.e.*, $\Phi$ will have $|\mathcal{S}|$ linearly independent columns). The $I$ matrix is another basis for $\mathbb{R}^{|\mathcal{S}|}$, yet no information flows between states (*i.e.*, the coverage of each feature is restricted to one state). When sorting columns of $\Phi_{\mathcal{F}_n}$ based on the size of the features, it starts with features with high coverage (excluding the null feature). As more conjunctions are introduced, the coverage is reduced exponentially (*i.e.*, the number of active features are decreased exponentially by the size of the feature set). Next, we explain how adding binary features can lead to guaranteed approximation error-bound reduction. We begin with a geometric Lemma used in Theorem 3.4.

**Lemma 3.3** *Let L be the plane specified by three distinct points $P, Q, C \in \mathbb{R}^d$, with $\alpha = \angle(CQ, CP) > 0$. Assume that the additional point $X \in \mathbb{R}^d$ is not necessarily in L. Define the angles $\beta = \angle(CX, CQ)$ and $\omega = \angle(CX, CP)$. Now let $P'$ denote the orthogonal projection of $P$ on $CX$. If $\alpha + \beta < \frac{\pi}{2}$, then $\|PP'\|$ is maximized when $CX \in L$.*

**Proof** In Appendix.

We now extend Theorem 3.6 of [Parr *et al.*, 2007] by deriving a lower bound ($\zeta x$) on the improvement caused by adding a new feature.

**Theorem 3.4** *Given an MDP with a fixed policy, where the value function is approximated as $\tilde{V}$, define $\delta = T(\tilde{V}) - \tilde{V}$, and $\|V - \tilde{V}\| = x > 0$, where $V$ is the optimal value for all states. Then $\forall \phi_f \in \mathbb{R}^{|\mathcal{S}|} : \beta = \angle(\phi_f, \delta) < \arccos(\gamma)$*

$$\exists \xi \in \mathbb{R} : \|V - \tilde{V}\| - \|V - (\tilde{V} + \xi\phi_f)\| \geq \zeta x, \quad (3)$$

*where $\gamma$ is the discount factor and*

$$\zeta = 1 - \gamma\cos(\beta) - \sqrt{1 - \gamma^2}\sin(\beta) < 1. \quad (4)$$

*Furthermore, if these conditions hold and $\tilde{V} = \Phi\theta$ with $\phi_f \notin \text{span}(\Phi)$ then:*

$$\|V - \Pi V\| - \|V - \Pi' V\| \geq \zeta x, \quad (5)$$

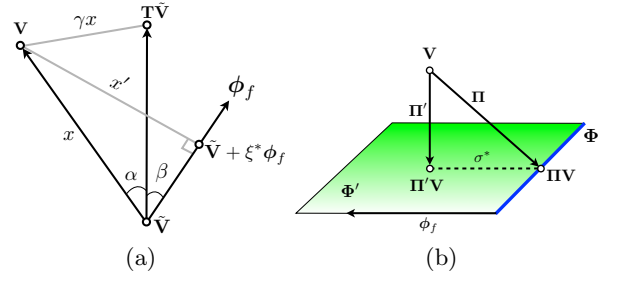

Figure 2: a) geometric view of $V, \tilde{V}, T(\tilde{V})$, and $\phi_f$. As $\beta$ shrinks $x'$ gets closer to $\gamma x$. b) increasing the dimensionality of the projection operator.

*where $\Pi$ and $\Pi'$ are orthogonal projection operators using $\Phi$ and $\Phi' = [\Phi \; \phi_f]$ respectively.*

**Proof** Consider both cases of the orientation of points $V$ and $T(\tilde{V})$ with respect to each other:

**Case $T(\tilde{V}) \neq V$** : Due to the contraction property of the Bellman operator, if $\|V - \tilde{V}\| = x$, then $\|V - T(\tilde{V})\| \leq \gamma x$. Define $\alpha$ as the $\angle(V - \tilde{V}, \delta)$, then using the sine rule:

$$\sin(\alpha) \leq \frac{\|V - T(\tilde{V})\|}{\|V - \tilde{V}\|} \leq \gamma \quad \Rightarrow \quad \alpha \leq \arcsin(\gamma)$$

Furthermore, by assumption, $0 \leq \beta < \arccos(\gamma) = \pi/2 - \arcsin(\gamma)$. Combined, these conditions indicate that $\alpha + \beta < \pi/2$.

For the next step, given $\xi > 0$, mapping the points $V, \tilde{V}, T(\tilde{V}), \tilde{V} + \xi\phi_f$ to $P, C, Q, X$ in Lemma 3.3 shows that the orthogonal projection length of vector $V - \tilde{V}$ on $\tilde{V} + \xi\phi_f - \tilde{V}$ is maximized when all four points are coplanar and $\omega = \angle(V - \tilde{V}, \xi\phi_f) = \alpha + \beta$. Notice that the coplanar argument is implicit in the proof of Theorem 3.6 of Parr *et al.* [2007]. Figure 2(a) depicts the geometric view in such a plane, where $\xi^* = \arg\min_\xi \|V - (\tilde{V} + \xi\phi_f)\|$, $x' = x\sin(\omega)$.[3] As shown above, $\alpha \leq \arcsin(\gamma)$ and $0 \leq \alpha + \beta < \frac{\pi}{2}$, thus $\sin(\alpha + \beta) \leq \sin(\arcsin\gamma + \beta) = \gamma\cos(\beta) + \sin(\beta)\sqrt{1 - \gamma^2}$, Hence,

$$x' \leq x\left(\gamma\cos(\beta) + \sqrt{1 - \gamma^2}\sin(\beta)\right)$$
$$x - x' \geq x\left(1 - \gamma\cos(\beta) - \sqrt{1 - \gamma^2}\sin(\beta)\right) \equiv \zeta x$$

Looking at Figure 2(a), it is easy to verify that $x - x' = \|V - \tilde{V}\| - \|V - (\tilde{V} + \xi\phi_f)\|$, which completes the proof for the case $T(\tilde{V}) \neq V$.

**Case $T(\tilde{V}) = V$** : This means that $\alpha = 0$. If $\phi_f$ crosses $V$, it means $\beta = 0$ and $\tilde{V} + \xi^*\phi_f = V$. Hence $\zeta = 1 - \gamma \cos(\beta) - \sqrt{1-\gamma^2} \sin(\beta) = 1 - \gamma$ and $\|\boldsymbol{V} - \tilde{\boldsymbol{V}}\| - \|\boldsymbol{V} - (\tilde{\boldsymbol{V}} + \xi^* \boldsymbol{\phi}_f)\| = \|\boldsymbol{V} - \tilde{\boldsymbol{V}}\| = x \geq \zeta x$. When $\boldsymbol{\phi}_f$ does not cross $\boldsymbol{V}$, together they form a plane in which $\|\boldsymbol{V} - \tilde{\boldsymbol{V}}\| - \|\boldsymbol{V} - (\tilde{\boldsymbol{V}} + \xi^* \boldsymbol{\phi}_f)\| = x(1 - \sin(\beta))$. In order to complete the proof, a lower bound for the above error reduction is derived:

$$0 < \beta < \arccos(\gamma) = \pi/2 - \arcsin(\gamma), 0 \leq \gamma < 1$$
$$\Rightarrow 0 < \beta + \arcsin(\gamma) < \pi/2$$

$$\Rightarrow \sin(\beta) \leq \sin(\beta + \arcsin(\gamma))$$
$$= \gamma \cos(\beta) + \sqrt{1-\gamma^2} \sin(\beta)$$

---
[3]Note that $\xi$ can take negative values as well, rendering $\phi_f$ important only as a line but not as a vector.

$(1 - \sin(\beta))x \geq (1 - \gamma \cos(\beta) - \sqrt{1-\gamma^2} \sin(\beta))x \equiv \zeta x$

Now we extend the proof to the linear function approximation case with $\tilde{\boldsymbol{V}} = \boldsymbol{\Phi}\boldsymbol{\theta}$. Since the first part of the proof holds for any approximation, let us consider the case where $\tilde{\boldsymbol{V}} = \boldsymbol{\Pi}\boldsymbol{V}$. Showing $\tilde{\boldsymbol{V}} + \xi^* \boldsymbol{\phi}_f = \boldsymbol{\Pi}'\boldsymbol{V}$ completes the proof as it turns Inequality 3 into Inequality 5. To proceed, we decompose $\boldsymbol{\phi}_f$ into two vectors $\boldsymbol{\phi}_f^{\parallel} \in \text{span}(\boldsymbol{\Phi})$ and $\boldsymbol{\phi}_f^{\perp} \perp \text{span}(\boldsymbol{\Phi})$. First, consider the case where $\boldsymbol{\phi}_f = \boldsymbol{\phi}_f^{\perp}$, Figure 2(b) provides a geometric view of the situation. The blue line and the green plane highlight $\text{span}(\boldsymbol{\Phi})$ and $\text{span}(\boldsymbol{\Phi}')$ respectively. Both $\boldsymbol{\Pi}$ and $\boldsymbol{\Pi}'$ are orthogonal projections into these subspaces. Hence, for any given value function $\boldsymbol{V}$, $\boldsymbol{\Pi}'\boldsymbol{V} = \boldsymbol{\Pi}\boldsymbol{V} + \sigma^* \boldsymbol{\phi}_f$, where, $\sigma^* \triangleq \arg\min_\sigma \|(\boldsymbol{V} - \boldsymbol{\Pi}\boldsymbol{V}) - \sigma \boldsymbol{\phi}_f\|$.

The extension to the case where $\boldsymbol{\phi}_f^{\parallel} \neq \boldsymbol{0}$ is straightforward. Consider a subspace defined by two representation matrices $\boldsymbol{\Phi}_1$ and $\boldsymbol{\Phi}_2$ (i.e., $\text{span}(\boldsymbol{\Phi}_1) = \text{span}(\boldsymbol{\Phi}_2)$), and corresponding orthogonal projection operators $\boldsymbol{\Pi}_1$ and $\boldsymbol{\Pi}_2$. Since both operators provide the solution to the same convex optimization (i.e., $\min_{\boldsymbol{\theta}} \|\boldsymbol{V} - \tilde{\boldsymbol{V}}\|$), where both domain and target space are identical, their outputs are equal (i.e., $\tilde{\boldsymbol{V}}_1 = \boldsymbol{\Pi}_1 \boldsymbol{V} = \boldsymbol{\Pi}_2 \boldsymbol{V} = \tilde{\boldsymbol{V}}_2$).[4] Hence if $\boldsymbol{\phi}_f^{\parallel}$ is added as the last column of $\boldsymbol{\Phi}'$, it does not change $\text{span}(\boldsymbol{\Phi}')$ and the result of the projection remains intact. ∎

The next theorem specializes the above result to the case of binary features, where new features are built using the conjunction operator.

**Theorem 3.5** *Given Assumptions A1-A3, $\boldsymbol{\chi} \subseteq \mathcal{F}_n, \tilde{\boldsymbol{V}} = \boldsymbol{\Phi}_{\boldsymbol{\chi}}\boldsymbol{\theta}, \boldsymbol{\delta} = \boldsymbol{T}(\tilde{\boldsymbol{V}}) - \tilde{\boldsymbol{V}}$, and $\|\boldsymbol{V} - \tilde{\boldsymbol{V}}\| = x > 0$, then $\forall f \in \text{pair}(\boldsymbol{\chi})$, if*

$$\eta_f = \frac{|\sum_{s \in \mathcal{S}, \phi_f(s)=1} \mathbf{d}(s)\boldsymbol{\delta}(s)|}{\sqrt{\left(\sum_{s \in \mathcal{S}, \phi_f(s)=1} \mathbf{d}(s)\right)\left(\sum_{s \in \mathcal{S}} \mathbf{d}(s)\boldsymbol{\delta}^2(s)\right)}} > \gamma$$

---
[4]While the corresponding coordinates, $\boldsymbol{\theta}$, in each case can be different, the resulting $\tilde{\boldsymbol{V}}$ are identical.

$$\exists \xi \in \mathbb{R} : \|\boldsymbol{V} - \tilde{\boldsymbol{V}}\| - \|\boldsymbol{V} - (\tilde{\boldsymbol{V}} + \xi\boldsymbol{\phi}_f)\| \geq \zeta x, \quad (6)$$
$$\|\boldsymbol{V} - \boldsymbol{\Pi}\boldsymbol{V}\| - \|\boldsymbol{V} - \boldsymbol{\Pi}'\boldsymbol{V}\| \geq \zeta x, \quad (7)$$
$$\text{where} \quad 1 - \gamma\eta_f - \sqrt{1-\gamma^2}\sqrt{1-\eta_f^2} = \zeta \quad (8)$$

**Proof** Theorem 3.4 provides a general rate of convergence for the error bound when arbitrary feature vectors are added to the feature matrix. Hence it is sufficient to show that the conditions of Theorem 3.4 holds in this new theorem, namely: 1) $\beta = \angle(\boldsymbol{\phi}_f, \boldsymbol{\delta}) < \arccos(\gamma)$ and 2) $\boldsymbol{\phi}_f \notin \text{span}(\boldsymbol{\Phi}_{\boldsymbol{\chi}})$. The latter is already shown through Theorem 3.2. As for the former:

$$\cos(\beta) = \frac{|\langle \boldsymbol{\phi}_f \cdot \boldsymbol{\delta} \rangle|}{\|\boldsymbol{\phi}_f\| \cdot \|\boldsymbol{\delta}\|} \quad (9)$$
$$= \frac{|\sum_{s \in \mathcal{S}, \phi_f(s)=1} \mathbf{d}(s)\boldsymbol{\delta}(s)|}{\sqrt{\left(\sum_{s \in \mathcal{S}, \phi_f(s)=1} \mathbf{d}(s)\right)\left(\sum_{s \in \mathcal{S}} \mathbf{d}(s)\boldsymbol{\delta}^2(s)\right)}}.$$

Therefore, $\beta = \arccos(\eta_f)$. By the assumption made earlier, $\eta_f > \gamma$. Hence $\beta < \arccos(\gamma)$. Satisfying the preconditions of Theorem 3.4, both Equations 3 and 5 are obtained. Switching $\cos(\beta)$ with $\eta_f$ in Equation 4 completes the proof. ∎

Theorem 3.5 provides sufficient conditions for a guaranteed rate of convergence in the error bound of the value function approximation by adding conjunctions of existing features. It leads directly to Theorem 3.1.

**Corollary 3.6** *An algorithm that selects features based on Equation 2, which maximizes Equation 9, is by our definition in Section 2.2 an MP algorithm.*

**Insight:** Equation 2 shows how feature coverage is a double-edged sword; while greater coverage includes more weighted Bellman error (i.e., the numerator) resulting in a higher convergence rate, it also contributes negatively to the rate of convergence (i.e., the denominator). The ideal feature would be active in a single state with all of the Bellman error. Intuitively, this conclusion is expected, because adding this ideal feature makes the approximation exact. When the weighted sum of Bellman errors is equal for a set of features, the feature with the least coverage is preferable. On the other hand, when all features have the same coverage, the one with the highest weighted Bellman error coverage is ideal. Another interesting observation is the relation between the difficulty of finding features that give the guaranteed error-bound convergence rate and the value of $\gamma$. In general, larger values of $\gamma$ render the MDP harder to solve. Here we can observe the same trend for finding good features as higher values of $\gamma$ reject more features in the set $\text{pair}(\boldsymbol{\chi})$ due to the constraint $\eta_f > \gamma$. Finally, we note that our theoretical results can be interpreted as a mathematical rationale for moving from a coarse to a fine representation, explaining empirical observations in both computer science [Whiteson *et al.*, 2007] and brain/cognitive science [Goodman *et al.*, 2008].

## 4 Empirical Results

Here we provide experimental evidence of Batch-iFDD's efficiency at policy evaluation, a crucial step in many reinforcement learning algorithms such as Policy Iteration. Policy evaluation is also the traditional setting for comparing feature expansion techniques [*e.g.,* Mannor and Precup, 2006; Painter-Wakefield and Parr, 2012]. We ran our experiments using the RLPy framework, which is available online [Geramifard *et al.*, 2013]. Results are presented in three classical RL domains: Mountain Car, Inverted Pendulum, and System Administrator. The last domain has more than one million states. Our ability to handle such a large domain is a direct consequence of the added efficiency and targeted search in Batch-iFDD.

On each iteration the best weights were found by running LSTD on $10^4$ samples gathered using the underlying policy. The $A$ matrix in LSTD was regularized by $10^{-6}$. Then the best feature was added to the representation using the corresponding expansion technique. All results were averaged over 30 runs and on each run all algorithms were exposed to the same set of samples. Standard errors are shown as shaded areas highlighting 95% confidence intervals.

We compared two approximations of Equation 2, shown in Equations 10 and 11. The first one comes from our theoretical analysis, where the steady state distribution is approximated by the collected samples. The second one is borrowed from previous work [Geramifard *et al.*, 2011]. In this section, we drop the implicit "Batch-" term and refer to these algorithms as iFDD$^+$ and iFDD[ICML-11] respectively.

$$\delta_i = r_i + [\gamma \phi(s'_i) - \phi(s_i)]^\top \boldsymbol{\theta}.$$

$$\tilde{f}_1^* = \underset{f \in \texttt{pair}(\chi)}{\operatorname{argmax}} \frac{|\sum_{i \in \{1,\cdots,m\}, \phi_f(s_i)=1} \delta_i|}{\sqrt{\sum_{i \in \{1,\cdots,m\}, \phi_f(s_i)=1} 1}}, \quad (10)$$

$$\tilde{f}_2^* = \underset{f \in \texttt{pair}(\chi)}{\operatorname{argmax}} \sum_{i \in \{1,\ldots,m\}, \phi_f(s_i)=1} |\delta_i| \quad (11)$$

We also implemented and compared against variants of OMP-TD [Painter-Wakefield and Parr, 2012], the previous state of the art MP algorithm in RL. Since this algorithm requires a set of potential features at initialization, we tested several different sizes of potential feature sets, each built by including features in increasingly finer levels of the concept lattice until the cap was reached. Note that in two dimensional problems (where only one layer of conjunction is available), if the OMP-TD potential feature pool contains every possible feature then the results of running OMP-TD and iFDD$^+$ will be identical, since both algorithms run the same optimization on the same set of features.

The first domain was Mountain Car [Sutton and Barto, 1998], with base features defined as discretizations of position and velocity into 20 partitions each, leading to 40 base features. The policy evaluated was to accelerate in the direction of the car's current velocity. Figure 3(a)-top shows the $\|\text{TD-Error}\|_2$ over the sample set for the iFDD and OMP-TD methods versus the number of feature-generating iterations. Since all the techniques start with the same set of features, their errors are identical at iteration zero. OMP-TD with pool sizes up to 250 did not capture the value function well. With $440 = 20 \times 20 + 40$ potential features, OMP-TD had access to all possible features, and as predicted it performed identically to iFDD$^+$. iFDD[ICML-11] performed similar to the best results. Figure 3(a)-bottom depicts the same results based on the wall-clock time. The iFDD techniques were at least 30 seconds faster than the OMP-TD methods as they considered fewer features on each iteration.

Next we considered the classical Inverted Pendulum domain [See Lagoudakis and Parr, 2003]. The feature setting was identical to the Mountain Car problem. The fixed policy pushed the pendulum in the opposite direction of its angular velocity. TD-error results as described before are presented for this domain in Figure 3(b). Again OMP-TD with 100 features did not have access to the necessary features and performed poorly. With 250 features, OMP-TD performed much better, but converged to a relatively less accurate representation after about 25 iterations. With access to the full set of features, OMP-TD(440) mirrored the performance of iFDD$^+$, both achieving the best results. In this domain, the less accurate approximation of Equation 2 used by iFDD[ICML-11] caused a significant drop in its performance compared to iFDD$^+$. However, it eventually exceeded the performance of OMP-TD(250) and caught up to iFDD$^+$ by expanding important features. There is a small initial rise in error for most of the algorithms, which we believe is due to the use of regularization. In terms of computational complexity, we see the same pattern as in the Mountain Car domain, where the iFDD methods were computationally more efficient (about 20% faster) than the OMP-TD techniques.

The third experiment considered the System Administrator domain [Guestrin *et al.*, 2001] with 20 computers and a fixed network topology. Each computer can either be up or down (following our example in Section 2.3), so there were 40 base features. The size of the state space is $2^{20}$. Computers go up or down depending on the status of their neighbors [See Guestrin *et al.*, 2001]. Allowed actions are to reboot one machine at each timestep or do nothing: in our case the policy uniformly randomly rebooted one of the down machines. Results are shown in Figure 3(c). The general theme remains the same, except for three observations: 1) for the first eight iterations, OMP-TD(2000) outperformed iFDD$^+$, 2) after 10 iterations iFDD$^+$ outperformed all of the OMP-TD techniques with pool sizes up to 2000, and 3) the speed advantage of the iFDD techniques was much more prominent. Based on clock time iFDD[ICML-11] was comparable to the best OMP-TD technique, while iFDD$^+$ achieved the final performance of OMP-TD(2000) more than 4 times faster. The reason for the initial OMP-TD success is that it was able to add complex features (conjunctions of several terms) early on without adding the coarser (subsumed) conjunctions that iFDD adds first.[5] The second observation is explained by the fact that iFDD$^+$ expands the set of potential features in a guided way, allowing it to discover crucial fine-grained features. Specifically iFDD$^+$ discovered features with 8 terms. Finally, as the size of the potential feature pool grew, the OMP-TD techniques required significantly more computation time. iFDD techniques on the other hand, only considered possible new pair-wise features, and scaled much better to larger MDPs.

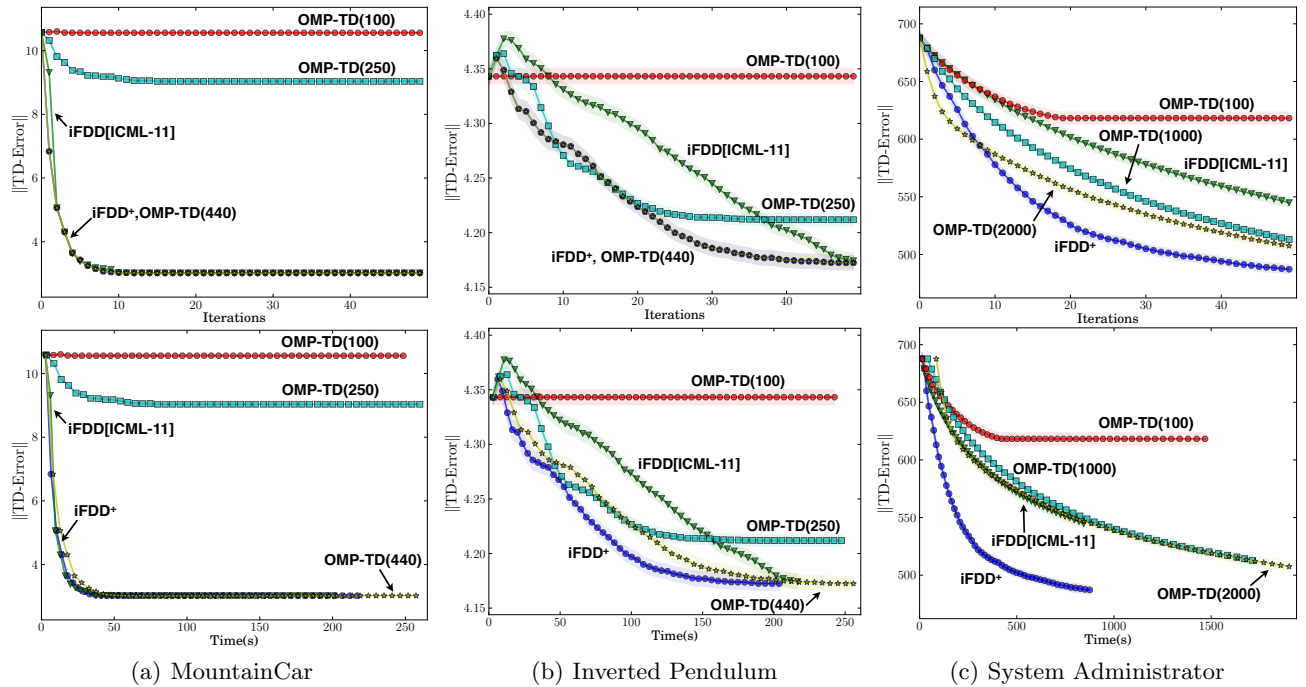

Figure 3: Empirical results from the a) Mountain Car, b) Inverted Pendulum, and c) System Administrator domains. The Y-axis depicts the $\ell_2$ norm of the TD prediction error plotted against (top) the number of feature expansions and (bottom) wall clock time. The colored shaded areas highlight the 95% confidence intervals.

## 5 Conclusions

We introduced Batch-iFDD (and Batch-iFDD$^+$) for feature construction in an RL setting and proved that it is a Matching Pursuit algorithm. Unlike previous MP techniques, Batch-iFDD expands the pool of potential features incrementally, hence searching the concept lattice more efficiently than previous MP techniques. Our empirical results support this finding as Batch-iFDD$^+$ outperformed the previous state of the art MP algorithm in three benchmarks, including a domain with over one million states.

It should be noted, that OMP-TD techniques are more general than Batch-iFDD techniques as they can work with arbitrary feature functions rather than binary functions. Also, Batch-iFDD is not immune to the poor selection of base features. For instance, in continuous state spaces base features can be built by discretizing each dimension using the indicator function, yet finding the "right" discretization for high dimensional problems can be challenging.

Beyond the results of this paper, Equation 2 provides insight as to why it is beneficial to add coarse features in the early stages of learning and finer features later on. In the early stages of learning, when feature weights have not been adjusted, the Bellman error is generally large everywhere. Therefore coarse features with large coverage have higher chances of having good error-bound convergence rates due to the numerator of Equation 2. As weights are updated, the Bellman error is reduced correspondingly. The reduced Bellman error will make the denominator of Equation 2 the deciding factor, rendering coarse features with large coverage ineffective. This transition may partially explain empirical results on RL agents exploring autonomously [Whiteson et al., 2007] and human subjects performing classification [Goodman et al., 2008], where both benefited from using coarse features at the beginning of learning but then progressed to finer-grained features to make better sense of a complex world.

---

[5]The feature pool for OMP-TD(2000) consisted of 760 and 1,200 features with 2 and 3 terms respectively.

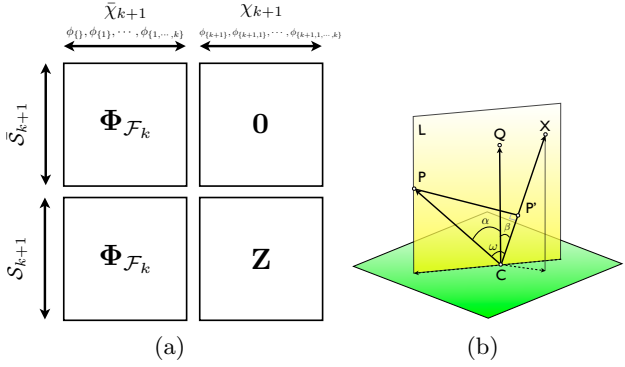

Figure 4: a) Depiction of $\mathbf{\Phi}_{\mathcal{F}_{k+1}}$ using $\mathbf{\Phi}_{\mathcal{F}_k}$ as the building block. Note that features are not sorted based on their cardinality order, but it does not change the rank of the resulting matrix. b) A $3D$ visualization of $d$ dimensional points $L$ and $Q$ and vector $X$ with $C$ as the center. $\|\boldsymbol{PP'}\|$ is maximized when $\omega = \alpha + \beta$.

## Acknowledgments

We thank Ronald Parr and our anonymous reviewers for insightful feedback. This research was sponsored by ONR grants N00014-07-1-0749 and N00014-11-1-0688.

## A Proof of Theorem 3.2

**Lemma A.1** *Given $m, n \in \mathbb{N}^+$ and $m \leq n$, if $\mathbf{X}_{m \times n}$ and $\mathbf{Z}_{m \times n}$ are full column rank matrices with real elements and $\mathbf{Y}_{m \times n}$ is arbitrary matrix, then matrix $\begin{bmatrix} \mathbf{X} & \mathbf{0} \\ \mathbf{Y} & \mathbf{Z} \end{bmatrix}$ is a full column rank matrix.*

**Proof** The proof follows from the definition of the matrix as both $\mathbf{X}$ and $\mathbf{Z}$ have linear independent columns.

**Theorem A.2** *Given Assumptions A1-A3, $\forall n \in \mathbb{N}^+$, $\mathbf{\Phi}_{\mathcal{F}_n}$ is invertible.*

**Proof** First note that $\mathbf{\Phi}_{\mathcal{F}_n}$ is a square matrix as $|\mathcal{F}_n| = |\mathcal{S}| = 2^n$. Hence it is sufficient to show that $\mathbf{\Phi}_{\mathcal{F}_n}$ has independent columns. The rest of the proof is through induction on $n$:

($n = 1$): The MDP has two states. Hence $\mathbf{\Phi}_{\mathcal{F}_1} = \begin{bmatrix} 1 & 0 \\ 0 & 1 \end{bmatrix}$, $det(\mathbf{\Phi}_{\mathcal{F}_1}) = 1$. Notice that the first column corresponds to the null feature (i.e., $\{\}$) which returns 1 for the single state with no active features.

($n = k$): Assume that $\mathbf{\Phi}_{\mathcal{F}_k}$ has independent columns.

($n = k+1$): Based on the previous assumption, $\mathbf{\Phi}_{\mathcal{F}_k}$ has linearly independent columns. Hence it is sufficient to show that $\mathbf{\Phi}_{\mathcal{F}_{k+1}}$ can be written as $\begin{bmatrix} \mathbf{X} & \mathbf{0} \\ \mathbf{Y} & \mathbf{Z} \end{bmatrix}$, where $\mathbf{X} = \mathbf{Y} = \mathbf{\Phi}_{\mathcal{F}_k}$, and $\mathbf{Z}$ has linearly independent columns. Lemma A.1 then completes the proof.

The new added dimension, $k + 1$, doubles the number of states because $|\mathcal{S}| = 2^{k+1}$. The new dimension also doubles the total number of possible features, as for any given set with size $k$, the total number of its subsets is $2^k$. Divide states into the two following sets: $\mathcal{S}_{k+1} = \{s|\phi_{\{k+1\}}(s) = 1\}, \bar{\mathcal{S}}_{k+1} = \{s|\phi_{\{k+1\}}(s) = 0\}$.

Similarly, divide features into two sets:
$$\chi_{k+1} = \{f|f \in \mathcal{F}_{k+1}, k+1 \in f\},$$
$$\bar{\chi}_{k+1} = \{f|f \in \mathcal{F}_{k+1}, k+1 \notin f\}.$$

Construct rows and columns of $\mathbf{\Phi}_{\mathcal{F}_{k+1}}$, following Figure 4. The values of the top left and bottom left of the matrix are $\mathbf{\Phi}_{\mathcal{F}_k}$, and the value of the top right of the matrix is $\mathbf{0}$. As for the bottom right ($\mathbf{Z}$), note that for all the corresponding states, $\phi_{\{k+1\}}(s) = 1$. Hence,
$$(\phi_{\{k+1\}}(s), \phi_{\{k+1,1\}}(s), \cdots, \phi_{\{k+1,1,\cdots,k\}}(s))$$
$$= (1, \phi_{\{1\}}(s), \cdots, \phi_{\{1,\cdots,k\}}(s)).$$

We know from the induction assumption that except for the first column, all other columns are linearly independent. Finally observe that the first column is the only column within $\mathbf{Z}$, with a 1 corresponding to the state with no active features and is independent of all other columns. ∎

Theorem 3.2 follows from Theorem A.2.

## B Proof of Lemma 3.3

**Proof** Let us first assume that $\boldsymbol{CX} \notin L$, hence there exists a three dimensional subspace defined by $\boldsymbol{CX}$ and $L$. Figure 4 depicts such a space. Then, $\mathrm{argmax}_\omega \|\boldsymbol{PP'}\| = \mathrm{argmax}_\omega \|\boldsymbol{CP}\| \sin(\omega) = \mathrm{argmax}_\omega \sin(\omega)$. Since $0 < |\alpha - \beta| \leq \omega \leq \alpha + \beta < \frac{\pi}{2}$, then $\mathrm{argmax}_\omega \|\boldsymbol{PP'}\| = \alpha + \beta$, which implies that $\boldsymbol{CX} \in L$ and thus is a contradiction. ∎

# References

Steven J. Bradtke and Andrew G. Barto. Linear least-squares algorithms for temporal difference learning. *Journal of Machine Learning Research (JMLR)*, 22:33–57, 1996.

Alborz Geramifard, Finale Doshi, Joshua Redding, Nicholas Roy, and Jonathan How. Online discovery of feature dependencies. In Lise Getoor and Tobias Scheffer, editors, *International Conference on Machine Learning (ICML)*, pages 881–888. ACM, June 2011.

Alborz Geramifard, Robert H Klein, and Jonathan P How. RLPy: The Reinforcement Learning Library for Education and Research. http://acl.mit.edu/RLPy, April 2013.

Noah D. Goodman, Joshua B. Tenenbaum, Thomas L. Griffiths, and Jacob Feldman. Compositionality in rational analysis: Grammar-based induction for concept learning. In M. Oaksford and N. Chater, editors, *The probabilistic mind: Prospects for Bayesian cognitive science*, 2008.

Carlos Guestrin, Daphne Koller, and Ronald Parr. Max-norm projections for factored mdps. In *International Joint Conference on Artificial Intelligence (IJCAI)*, pages 673–682, 2001.

Michail G. Lagoudakis and Ronald Parr. Least-squares policy iteration. *Journal of Machine Learning Research (JMLR)*, 4:1107–1149, 2003.

Stephen Lin and Robert Wright. Evolutionary tile coding: An automated state abstraction algorithm for reinforcement learning. In *AAAI Workshop: Abstraction, Reformulation, and Approximation*, Atlanta, Georgia, USA, 2010.

Sridhar Mahadevan, Mauro Maggioni, and Carlos Guestrin. Proto-value functions: A Laplacian framework for learning representation and control in Markov decision processes. *Journal of Machine Learning Research (JMLR)*, 8:2007, 2006.

Shie Mannor and Doina Precup. Automatic basis function construction for approximate dynamic programming and reinforcement learning. In *International Conference on Machine Learning (ICML)*, pages 449–456. ACM Press, 2006.

Christopher Painter-Wakefield and Ronald Parr. Greedy algorithms for sparse reinforcement learning. In *International Conference on Machine Learning (ICML)*, pages 968–975. ACM, 2012.

Ronald Parr, Christopher Painter-Wakefield, Lihong Li, and Michael Littman. Analyzing feature generation for value-function approximation. In *International Conference on Machine Learning (ICML)*, pages 737–744, New York, NY, USA, 2007. ACM.

Bohdana Ratitch and Doina Precup. Sparse distributed memories for on-line value-based reinforcement learning. In *European Conference on Machine Learning (ECML)*, pages 347–358, 2004.

David Silver, Richard S. Sutton, and Martin Müller. Sample-based learning and search with permanent and transient memories. In *International Conference on Machine Learning (ICML)*, pages 968–975, New York, NY, USA, 2008. ACM.

Peter Stone, Richard S. Sutton, and Gregory Kuhlmann. Reinforcement learning for RoboCup-soccer keepaway. *International Society for Adaptive Behavior*, 13(3):165–188, 2005.

Nathan R. Sturtevant and Adam M. White. Feature construction for reinforcement learning in hearts. In *5th International Conference on Computers and Games*, 2006.

Richard S. Sutton and Andrew G. Barto. *Reinforcement Learning: An Introduction*. MIT Press, 1998.

Richard S. Sutton. Learning to predict by the methods of temporal differences. *Machine Learning*, 3:9–44, 1988.

Shimon Whiteson, Matthew E. Taylor, and Peter Stone. Adaptive tile coding for value function approximation. Technical Report AI-TR-07-339, University of Texas at Austin, 2007.